\documentclass[letterpaper, 10pt, journal, twoside]{IEEEtran}

\usepackage{amsmath}
\usepackage{amssymb}
\usepackage{graphicx}
\usepackage[dvipsnames]{xcolor}
\usepackage[bookmarks=true,colorlinks=true]{hyperref}
\usepackage{comment}
\usepackage{booktabs}
\usepackage{makecell}
\usepackage{multirow}
\usepackage{cite}
\usepackage{enumitem}

\def\BibTeX{{\rm B\kern-.05em{\sc i\kern-.025em b}\kern-.08em
    T\kern-.1667em\lower.7ex\hbox{E}\kern-.125emX}}
\usepackage{balance}

\usepackage[export]{adjustbox}
\usepackage[normalem]{ulem}
\useunder{\uline}{\ul}{}

\newcommand\scalemath[2]{\scalebox{#1}{\mbox{\ensuremath{\displaystyle #2}}}}
\newcommand{\method}{ZeST}

\begin{document}
\title{\method{}: A VLM-based Zero-Shot Traversability Navigation for Unknown Environments}
\author{Shreya Gummadi$^{1\dagger}$, Mateus V. Gasparino$^{1\dagger}$, Gianluca Capezzuto$^{2\dagger}$, Marcelo Becker$^{2}$, and Girish Chowdhary$^{1}$
\thanks{Manuscript received: June 29, 2026; Accepted: October 1, 2026.}%
\thanks{This paper was recommended for publication by Editor Giuseppe Loianno upon evaluation of the Associate Editor and Reviewers' comments.}%
\thanks{This work was supported in part by CNPq \#308092/2020-1, AIFARMS \#1024178, NSF-USDA COALESCE \#2021-67021-34418, and iFARM \#2022-77038-37306. G. Capezzuto was funded by the S\~ao Paulo Research Foundation (FAPESP), \#2024/00519-0.}%
\thanks{$\dagger$ These authors contributed equally to this work.}%
\thanks{$^{1}$Field Robotics Engineering and Science Hub (FRESH), Illinois Autonomous Farm, University of Illinois at Urbana-Champaign (UIUC), IL.}%
\thanks{$^{2}$Mobile Robotics Group, University of S\~ao Paulo, S\~ao Carlos, SP, Brazil.}%
\thanks{The publication was written prior to Mateus V. Gasparino joining Amazon.}%
\thanks{Correspondence to: {\tt\footnotesize \{gummadi4,girishc\}@illinois.edu}}%
\thanks{Source code is available at: \href{https://github.com/matval/zest_navigation}{\nolinkurl{github.com/matval/zest_navigation}}}%
\thanks{Digital Object Identifier (DOI): see top of this page.}%
}
\markboth{IEEE Robotics and Automation Letters. Preprint Version. Accepted October 2026}
{Gummadi \MakeLowercase{\textit{et al.}}: ZeST: A VLM-based Zero-Shot Traversability Navigation for Unknown Environments}

\maketitle

\begin{abstract}

Advances in robotics and autonomous navigation systems hinge on the ability to predict terrain traversability accurately. Traditional methods for generating datasets to train these prediction models often involve putting robots into potentially hazardous environments, posing risks to equipment and safety. To solve this problem, we present \method{}, a novel approach that treats repeated VLM outputs as stochastic measurements and fuses them into an uncertainty-aware posterior. Our approach not only performs zero-shot traversability and mitigates the risks associated with real-world data collection but also accelerates the development of advanced navigation systems, offering a cost-effective and scalable solution. To support our findings, we present navigation results, in both controlled indoor and unstructured outdoor environments. As shown in the experiments, \method{} provides safer navigation with a 90--100\% success rate while no baseline exceeds 60\%, and keeps complete success under query latencies up to $3.6\,\text{s}$.

\end{abstract}


\section{Introduction}

\IEEEPARstart{T}{he} development of autonomous navigation systems is a cornerstone of robotics, with terrain traversability prediction being a critical component~\cite{gasparino2022wayfast,gasparino2024wayfaster,gummadi2024fed,cai2023evora,frey2023fast}. Robots must determine whether a given terrain is passable or poses risks to their operation in applications such as planetary exploration, disaster response, and autonomous vehicles. However, accurately predicting terrain traversability remains challenging due to the diversity and unpredictability of real-world environments~\cite{gasparino2022wayfast}.

Reliably capturing traversability requires more than a single sensing cue or fixed rule~\cite{borges2022survey}. Geometry alone is insufficient, since flat-looking terrain can still be hazardous, and compliant vegetation such as tall grass or foliage may remain traversable~\cite{sevastopoulos2022survey}. Semantic understanding can help disambiguate geometric cases, but the mapping from semantic class to traversability cost is itself context-dependent and varies across environments~\cite{cai2022speeddist}. With advances in foundation models, broad zero-shot priors for navigation have become available~\cite{huang2023visual, sathyamoorthy2024convoi}. However, their raw outputs are not inherently calibrated and can vary from query to query, so a single prediction cannot consistently support safety-critical decisions~\cite{levine2023calibration, tu2024calibratingvlms}.

Although Visual-Language Model (VLM) outputs exhibit response variability, repeated estimates for a fixed terrain region can still provide informative measurements of underlying traversability. Building on this observation, we reinterpret the usage of VLMs not as a deterministic oracle, but as a stochastic measurement model: each query produces a traversability sample, and aggregating multiple predictions defines both an estimate of terrain cost and the uncertainty associated with that assessment. Using a multimodal foundation model promises the navigation system access to broad contextual priors, while casting the VLM outputs as stochastic measurements converts the predictions into uncertainty-aware navigation constraints.
\begin{figure}[t]
    \centering
    \frame{\includegraphics[trim={0 2cm 0 2cm},clip,width=1.0\linewidth]{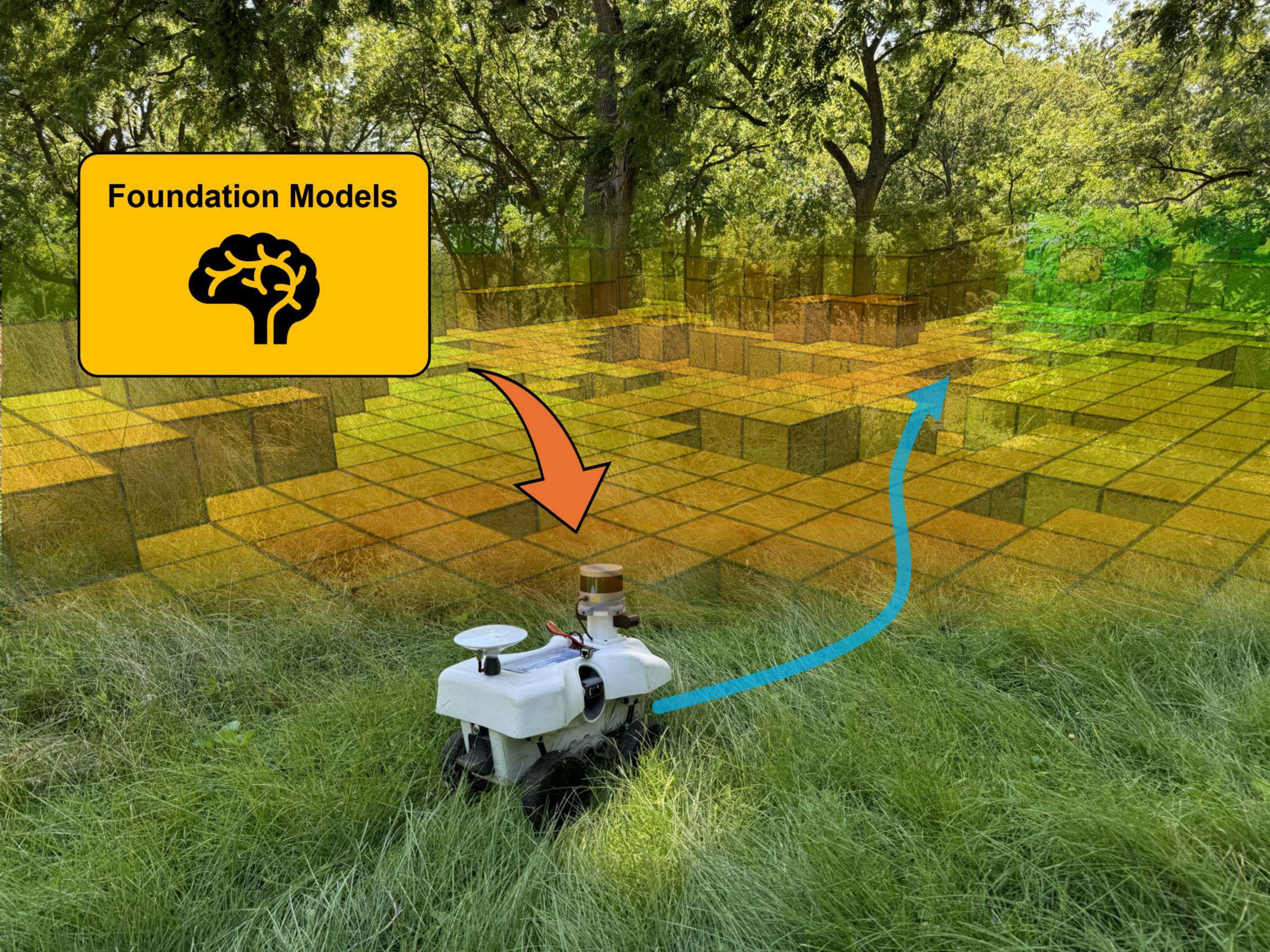}}
    \caption{\textbf{\method{}.} A zero-shot navigation framework that treats a multimodal foundation model as a stochastic traversability sensor, fuses visual-language predictions into an uncertainty-aware map, and plans risk-aware trajectories.}
    \label{fig:main-figure}
    \vspace{-0.25in}
\end{figure}

\begin{figure*}[t]
    \centering
    \includegraphics[width=\linewidth]{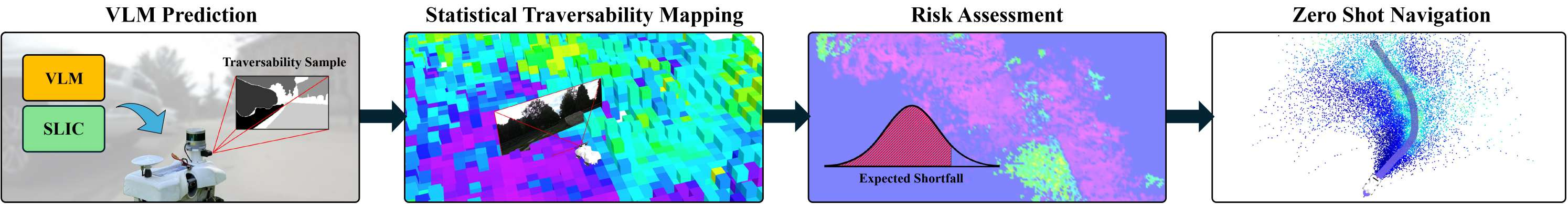}
    \caption{\textbf{\method{} System Overview}. (1) \textbf{VLM Prediction}: SLIC-based numbered image used by VLM to generate traversability samples. (2) \textbf{Statistical Traversability Mapping}: Samples are fused into a probabilistic map. (3) \textbf{Risk Assessment}: Navigation risk is quantified via expected shortfall. (4) \textbf{Zero-Shot Navigation}: The robot navigates unfamiliar environments using the risk-aware map.}
    \label{fig:main-algorithm}
    \vspace{-0.2in}
\end{figure*}
    
In this work, we propose \textbf{\method{}} (\textbf{Ze}ro-\textbf{S}hot \textbf{T}raversability), a novel framework that decouples asynchronous VLM inference from high-frequency traversability predictions, as shown in Fig.~\ref{fig:main-figure}. To address prediction variability inherent in VLMs, \method{} models traversability with a Normal Inverse Gamma (NIG) distribution over the mean and variance of repeated VLM outputs, explicitly capturing both aleatoric and epistemic uncertainties. With the NIG model, \method{} builds coherent global maps of navigational costs while quantifying uncertainty. Our contributions are:

\begin{itemize}
    \item An online Bayesian fusion model that fuses VLM traversability outputs into a single uncertainty-aware posterior, recovering per-region cost estimate and uncertainty through closed-form conjugate updates;
    \item A risk-aware planner built on the traversability posterior that accounts for worst-case cost through a Conditional Value-at-Risk (CVaR) objective and modulates robot speed by epistemic uncertainty, slowing in under-observed regions while running in real time onboard;
    \item Comprehensive real-world evaluations in challenging indoor and outdoor scenarios, demonstrating improved field robot navigation over conventional methods through a fully zero-shot deployment with no in-domain training.
\end{itemize}

\section{Related Work}

\paragraph{Traversability Prediction} Traditional approaches to traversability prediction often rely on heuristic-based methods~\cite{wermelinger2016navigation, borges2022survey, sevastopoulos2022survey}. While effective in structured or well-defined settings, such methods depend on hand-designed rules and often struggle to generalize across diverse terrains. Another common approach is to learn traversability predictors from labeled data using supervised models~\cite{sivakumar2021learned,sivakumar2024lessons}, which can improve prediction accuracy but require costly manual annotation. More recent methods instead learn from robot experience through self-supervision~\cite{kahn2021badgr, gasparino2022wayfast, gasparino2024wayfaster, frey2023fast}, reducing labeling effort but requiring in-domain interaction data and adaptation to the deployment environment. Simulation-based methods further reduce real-world training risk by learning traversability from physics-based rollouts or robot-specific models~\cite{Frey2022LocomotionPG, Lee2024LearningRA}, but they still depend on platform-specific training pipelines, and transfer across domains can be challenging. In contrast, \method{} performs zero-shot traversability directly from visual observations at deployment time.

\paragraph{Foundation Models for Robot Navigation} A growing body of work uses language and vision-language models as priors for navigation~\cite{shah2023lm, chen2023a2, huang2023visual, sathyamoorthy2024convoi}. Several methods parse free-form goals or instructions into subgoals that a downstream policy executes~\cite{shah2023lm, chen2023a2}, while others project VLM features into spatial maps for language-grounded queries~\cite{huang2023visual}. A recurring obstacle is cost. VLM inference can take seconds and exceed typical onboard compute, which makes a single high-rate prediction unsuitable for safety-critical control. CoNVOI~\cite{sathyamoorthy2024convoi} sidesteps latency by querying the model only to rank precomputed path proposals and performing classical LiDAR-based obstacle avoidance, but treats each VLM output as reliable and discards the prediction noise that arises when the same scene is queried repeatedly. \method{} instead keeps that noise and treats repeated VLM outputs as stochastic measurements to be fused rather than trusted individually.

\paragraph{Probabilistic Traversability Mapping} Prior work also models uncertainty for traversability and risk-aware planning~\cite{fan2021riskaware, endo2023riskaware, cai2023evora, fan2021step, triest2024unrealnet}. Such methods learn uncertainty-aware traversability from task-specific models and in-domain experience, then plan conservatively over the learned estimates. EVORA~\cite{cai2023evora}, for example, learns a traction distribution that separates aleatoric from epistemic uncertainty for CVaR-based risk-aware planning. A property shared across these methods is that the uncertainty model is trained offline from in-domain interaction, so the traversability sensor itself must be relearned for each deployment environment. \method{} draws instead on Bayesian occupancy mapping, where conventional sensors update a per-cell posterior through conjugate counting updates~\cite{doherty2019learning, gan2020bayesian, kim2024evidential}. Such maps were designed for geometric or semantic occupancy, with binary or class-valued cells and clean sensor readings. \method{} extends the counting-sensor formulation to continuous traversability, replacing the conventional sensor with a zero-shot VLM and treating its outputs as stochastic measurements in an online posterior. \method{} therefore requires no learned sensor and no in-domain data, taking the pretrained foundation model as given and placing the modeling effort in online fusion for risk-aware planning.

\section{System Design}

\method{}, illustrated in Fig.~\ref{fig:main-algorithm}, is a modular navigation system that estimates traversability in unstructured, unknown environments without prior knowledge or in-domain data collection by integrating VLM predictions with a probabilistic mapping and planning framework.

\begin{figure}[htp]
    \centering
    \frame{\includegraphics[trim={0 0cm 0 0cm},clip,width=1.0\linewidth]{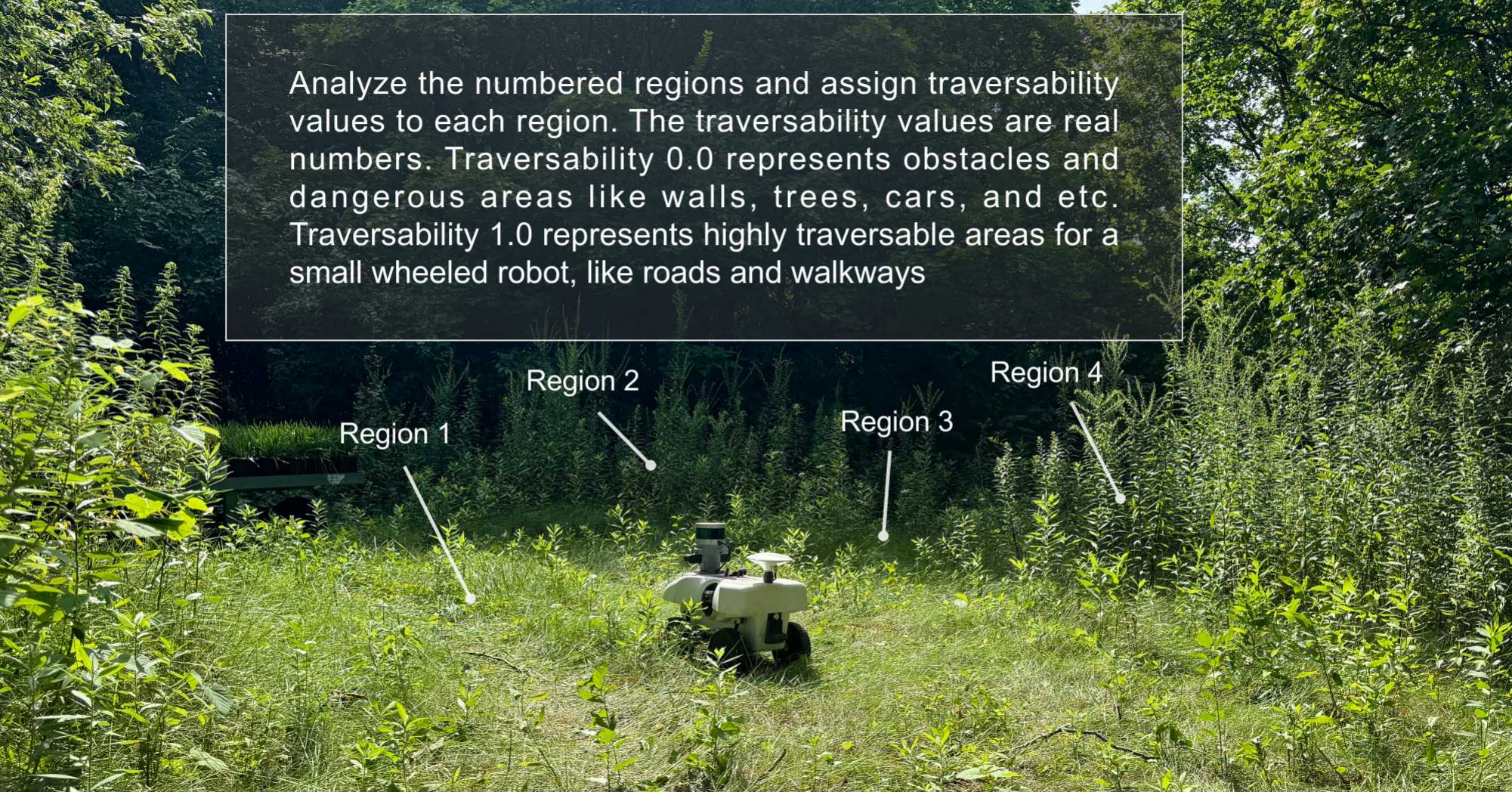}}
    \caption{\textbf{VLM Traversability Query.} A structured prompt combining physical robot characteristics, numeric terrain anchors, and segmented image regions to condition the foundation model for continuous traversability outputs.}
    \label{fig:zest-prompt}
    \vspace{-0.1in}
\end{figure}

\subsection{VLM Prediction}

\subsubsection{Mask Generation} 

Since VLMs do not explicitly predict traversability, we add a pre-processing step that segments the input image so the VLM can be queried region by region. Given an RGB image, we use the off-the-shelf Simple Linear Iterative Clustering (SLIC)~\cite{achanta2012slic} to generate masks. The segmentation is based on visual similarity, assuming that pixels with comparable features have similar traversability values. Each mask is overlaid on the image with a unique identifier from $1$ to $M$, the number of masks, so the VLM returns region-specific predictions.

\subsubsection{Querying the Visual-Language Model}

To analyze the segmented regions of an image, we leverage a VLM to provide continuous traversability scores for each region. As shown in Fig.~\ref{fig:zest-prompt}, the prompt specifies the robot's size and mobility characteristics and provides example terrains (e.g., grass, flat terrain, and obstacles) with reference traversability values, conditioning the VLM to infer the output scale and reason about how terrain features affect traversability for the given robot. The same prompt is used across environments, so output variations arise from visual differences rather than prompt formulation. The prompt is paired with the numbered image from the previous step, and the VLM outputs a list of traversability values ordered by region index.

\subsection{Learning a Traversability Distribution}

VLM traversability scores can vary across repeated queries of the same scene and across environmental conditions due to the inherent uncertainty in the model's reasoning. We therefore model traversability as a latent distribution rather than a single deterministic value.

Consider the robot queries the VLM $n \in \mathbb{N}$ times, obtaining $n$ measurements $X = \{x_1, x_2, \dots, x_n\}$. We assume that traversability for each region can be modeled as a Gaussian distribution. However, each query only returns traversability values sampled from the true underlying distribution:
\begin{equation} \label{eq:vlm-sampling-model}
   x_i \sim \mathcal{N}(m_i, \sigma_i^2), \quad i = 1, \dots, n,
\end{equation}
where $m_i$ and $\sigma_i^2$ represent the unknown mean and variance of the Gaussian distribution for region $i$. The goal is to estimate these parameters, $m_i$ and $\sigma_i$, which define the latent distribution of traversability for each region.

Our approach draws inspiration from the counting sensor model for occupancy grid mapping, as outlined in~\cite{hahnel2003map} and further explored in~\cite{thrun2003learning, doherty2019learning, gan2020bayesian, kim2024evidential}. The counting sensor model operates on a discretized grid of map cells, indexed by $j \in \mathbb{Z}^+$, and describes the occupancy probability $\theta_j$ for each cell using a Bernoulli likelihood:
\begin{equation}
    p(y_i|\theta_j) = \theta_j^{y_i} (1-\theta_j)^{1-y_i},
\end{equation}
where $Y = \{y_1, ..., y_N\}$ represents observations, with $y_i \in \{0, 1\}$ indicating whether a beam was reflected by ($y_i = 1$) or passed through ($y_i = 0$) the map cell at beam endpoint $\mathbf{b}_i$. The beam endpoints are given by $B = \{\mathbf{b}_1, \ldots, \mathbf{b}_N \mid \mathbf{b}_i \in \mathbb{R}^3\}$. Each range measurement consists of a pair $(\mathbf{b}_i, y_i)$, and the goal is to compute the posterior distribution of the occupancy probability $\theta_j$ given the observations $B$ and $Y$.

Using a conjugate Beta prior, $\mathrm{Beta}(\alpha_0,\beta_0)$, where $\alpha_0$ and $\beta_0$ are prior hyperparameters typically chosen as small, uninformative values, we can compute the posterior distribution over $\theta_j$. The posterior is given by:
\begin{equation}
        p(\theta_j | B,Y) = \mathrm{Beta}(\alpha_j,\beta_j),
\end{equation}
where $\alpha_j$ and $\beta_j$ are updated based on the observations. This formulation, widely used in robotic occupancy-grid mapping~\cite{doherty2019learning}, allows for efficient Bayesian updates.

\begin{figure}[!t]
    \centering
    \includegraphics[width=1.0\linewidth]{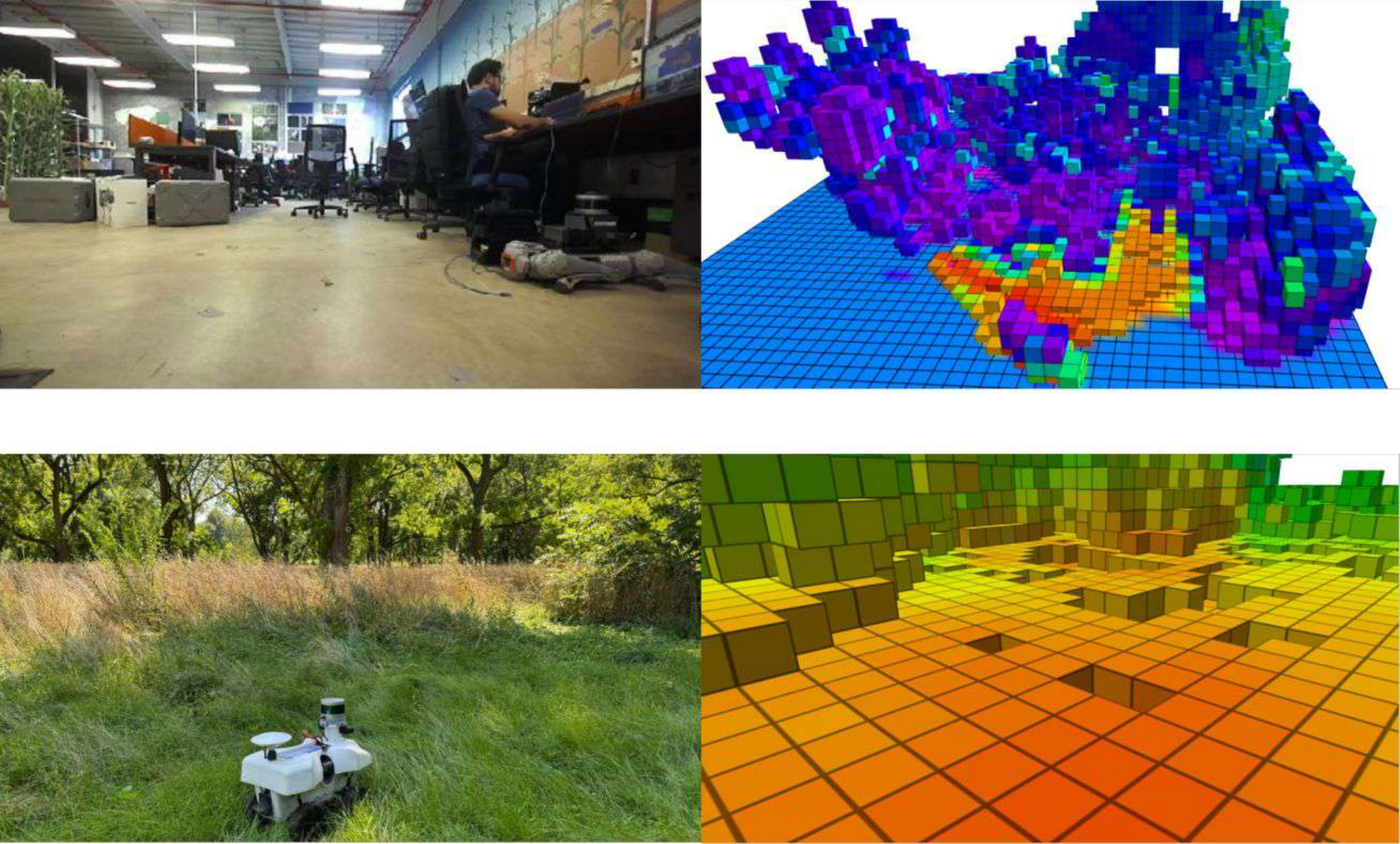}
    \caption{\textbf{Traversability Map Examples.} Qualitative examples of 3D traversability maps generated by \method{}. Each row shows the RGB observation on the left and the corresponding traversability map on the right. (Top) \textbf{Indoor Environment:} \method{} identifies the open floor region as traversable while assigning lower traversability to cluttered areas around desks, chairs, and obstacles. (Bottom) \textbf{Outdoor Environment:} \method{} produces a traversability map for a grassy scene, highlighting feasible regions through dense vegetation and uneven terrain. Warmer colors represent more traversable regions.}
    \label{fig:zest-octomap-example}
\end{figure}

As shown in Fig.~\ref{fig:zest-octomap-example}, we extend this model beyond occupancy estimation by integrating it with traversability assessment. Specifically, we estimate traversability values using an NIG distribution. The NIG distribution provides a flexible framework for handling uncertainty in traversability assessments, particularly in unstructured outdoor environments where sensor noise and terrain variations are prevalent~\cite{wermelinger2016navigation}.

For each cell, we model traversability as a random variable whose posterior parameters are governed by repeated VLM measurements, separating aleatoric uncertainty (inherent measurement noise) from epistemic uncertainty (limited observations). Representing both components is essential for safe navigation in challenging conditions.

The NIG distribution is a conjugate prior for the mean $m$ and variance $\sigma^2$ of a Gaussian (Normal) distribution. As a conjugate prior, the posterior distribution of $m$ and $\sigma^2$, given observed data, remains within the NIG family. Conditioned on the observations in $X$, the posterior parameters are given by:

\resizebox{0.9\linewidth}{!}{%
\begin{minipage}{\linewidth}
    \begin{align*}
        &\gamma_n = \frac{\kappa_0 \gamma_0 + n \bar{x}}{\kappa_n}, \quad &&\kappa_n = \kappa_0 + n, \\
        &a_n = a_0 + \frac{n}{2}, &&b_n = b_0 + \frac{1}{2}\sum_{i=1}^n (x_i-\bar{x})^2 + 
        \frac{\kappa_0 n (\bar{x}-\gamma_0)^2}{2(\kappa_0+n)},
    \end{align*}
\end{minipage}%
}
\vspace{0.1in}

\noindent where $\bar{x}$ is the sample mean of $X$. For cells with no associated VLM measurements ($n = 0$), the NIG prior $(\gamma_0,\kappa_0,a_0,b_0)$ is retained, combining prior knowledge with VLM observations into a single per-region posterior over traversability.

\subsection{Risk Assessment}

To assess the risk of the robot navigating into a given space, we compute the marginal distribution of $m$, which is represented as a Student's t-distribution~\cite{montgomery2010applied}, where the parameters of the Marginal Student's t-Distribution are:
\begin{equation*}
    \scalemath{0.85}{
        \underbrace{2a_n}_\text{Degrees of Freedom} \qquad
        \underbrace{\gamma_n}_\text{Location (Mean)} \qquad
        \underbrace{s = \sqrt{\frac{b_n (\kappa_n+1)}{a_n \kappa_n}}}_\text{Scale}.
    }
\end{equation*}

In the case of \method{}, the traversability map is given as an OctoMap. Therefore, to compute the CVaR, we take the worst-case scenario within each pillar of the map, compressing the z-axis dimension and generating a grid map. Since low traversability values correspond to higher risk, we summarize each cell by the expected traversability over the least-traversable portion of its predictive distribution. We fix a fraction $c \in (0,1)$ and average $m$ over its lower $c$-fraction, with $c=0.95$ throughout. The Value at Risk (VaR) at level $c$ is the $c$-quantile of $m$:
\begin{equation} \label{eq:var}
  \text{VaR}_c = \gamma_n + s \cdot t_{2a_n,c},
\end{equation}
where $t_{2a_n,c}$ denotes the $c$-quantile of the standard Student's t-distribution with $2a_n$ degrees of freedom, i.e., $F(t_{2a_n,c}) = c$.

The CVaR$_{c}$ is then the conditional mean of $m$ below this quantile:
\begin{equation} \label{eq:cvar-def}
    \scalemath{0.85}{
        \text{CVaR}_c = \mathbb{E}[m~|~m \leq \text{VaR}_c].
    }
\end{equation}

For a Student's t-distribution, CVaR can be computed using:
\begin{equation} \label{eq:cvar-closed}
    \scalemath{0.85}{
        \text{CVaR}_c = \gamma_n - s \cdot
        \frac{2a_n + (t_{2a_n,c})^2}{2a_n - 1}
        \cdot
        \frac{f(t_{2a_n,c})}{c},
    }
\end{equation}
where $f(t_{2a_n,c})$ is the Probability Density Function (PDF) of the standard Student's t-distribution evaluated at $t_{2a_n,c}$, and the denominator $c = F(t_{2a_n,c})$ is the probability mass of the averaged region. Taking $c$ close to one yields mild mass-based risk aversion, discounting cells with a non-negligible probability of low traversability while preserving score spread.

To avoid costly quantile computation at runtime, we precompute quantiles over a range of degrees of freedom $\nu = 2a_n$ and interpolate between stored values with cubic splines. Since the t-distribution converges to the normal distribution as $2a_n \rightarrow \infty$~\cite{montgomery2010applied}, we switch to the normal distribution for both quantile and PDF evaluation when $2a_n \geq 40$. Together, the two approximations speed up the risk computation $50\times$ on a Jetson AGX Orin with minimal loss of accuracy.

\subsection{Traversability-based Path Planning} 
\label{subsec:zest-path-planning}

Given the robot's current state, the target pose, and the traversability map, we plan a short, feasible path to the goal. We use the sampling-based Rapidly-exploring Random Tree$^*$ (RRT$^*$) algorithm~\cite{karaman2011anytime} for rapid planning across long distances, replacing traditional collision checks with our CVaR cost to guide the planner.

We approximate the robot footprint as a rectangle aligned with its orientation, with dimensions equal to half the robot's size, in pose $(x,y,\psi)$. The traversability metric for each pose is the average local CVaR of all cells covered by the footprint.

For traversability-based path planning, we adapt RRT$^*$ to score candidate edges using terrain risk and prediction uncertainty, rather than collision-free connectivity alone. The edge cost between adjacent poses $(x, y, \psi)$ and $(x', y', \psi')$ combines $c_{\text{CVaR}}$ and $c_\kappa$, where $c_{\text{CVaR}}$ captures traversability risk and $c_\kappa$ penalizes epistemic uncertainty. The traversability cost is:
\begin{equation} \label{eq:c-cvar}
    c_{\text{CVaR}} = e^{-w_{\text{trav}} \cdot \text{CVaR}(\mathbf{p})},
\end{equation}
where $w_{\text{trav}}$ controls how sharply low-CVaR cells are penalized and $\text{CVaR}(\mathbf{p})$ denotes the CVaR value associated with pose $\mathbf{p} = (x, y, \psi)$, averaged over all grid cells covered by the robot's footprint. The uncertainty cost favors frequently observed areas:
\begin{equation} \label{eq:c-kappa}
    c_{\kappa} = w_{\text{cov}} \cdot e^{\kappa_b - \kappa(\mathbf{p})},
\end{equation}
where $w_{\text{cov}}$ scales the cost of under-observed cells, and $\kappa(\mathbf{p})$ is the average observation count within the robot footprint at pose $\mathbf{p}$. The bias $\kappa_b$ keeps the exponential term properly scaled, so low-confidence path segments receive higher cost.

We deliberately omit an explicit distance cost: longer paths inherently accumulate higher traversability and uncertainty costs, so the objective implicitly favors shorter routes.

\subsection{Traversability-based Model Predictive Controller}

The RRT$^*$ planner of Section~\ref{subsec:zest-path-planning} produces a global reference path over the traversability-risk map, which is then tracked locally by Model Predictive Path Integral (MPPI) control~\cite{williams2017information} with horizon $N \in \mathbb{N}$. MPPI reads CVaR values and counts through the same footprint approximation, samples random action sequences, and solves:
\begin{align} \label{eq:zest-optimization}
    \scalemath{0.85}{
        \begin{gathered}
            \min_{u_{k:k+N-1}}
            \sum_{i=k}^{k+N-1} \left( ||\mathbf{x}_i - \mathbf{x}_i^r||^2_Q + \mathcal{J}_{i,u} \right) \\
            + ||\mathbf{x}_{k+N} - \mathbf{x}_{k+N}^r||^2_{Q_N} - \sum_{i=k}^{k+N} w_{\text{safe}} \cdot \text{CVaR}(\mathbf{x}_i),
        \end{gathered}
    }
\end{align}
where the first term minimizes the error between the predicted states $\mathbf{x}_i$ and the reference states $\mathbf{x}_i^r$ along the path, ensuring accurate tracking; the second term penalizes deviations from the reference actions; the third term minimizes the terminal cost, driving the robot toward the goal location; and the fourth term maximizes CVaR over traversability predictions to prioritize safety.

For \method{}, we design a speed-conditioned epistemic uncertainty cost, where the desired speed is changed according to the uncertainty attributed to that predictive path. Since traversability is predicted using a counting model, as more measurements are gathered from a space, the epistemic uncertainty in that corresponding area will be lower. Therefore, we want the robot to navigate more slowly in areas where uncertainty is high, so that it has more time to learn the true distribution of that location. We define the action cost as: \begin{equation} \label{eq:action-cost}
    \mathcal{J}_{i,u} = ||u_i - u_i^r \cdot (1-e^{(\kappa_b - \kappa(\mathbf{x}_i)) \cdot w_{\text{slow}}})||^2_R,
\end{equation}
where $\kappa_b$ and $\kappa(\cdot)$ carry the same meaning as in Section~\ref{subsec:zest-path-planning}.

$Q$, $Q_N$, and $R$ are weight matrices; $w_{\text{safe}}$ weights the CVaR safety term and $w_{\text{slow}}$ sets how aggressively the controller reduces speed in uncertain cells. After minimizing the cost function, the first command in the control horizon is used to drive the robot towards the goal.

\section{Experimental Results}

In this section, we evaluate \method{} across different real-world environments, ranging from cluttered indoor spaces to unstructured outdoor terrains. We aim to demonstrate the system's ability to generalize to unseen scenarios in a zero-shot manner, while validating its capability for safer navigation through comparisons against baselines. 

\subsection{Implementation Details}

We deploy \method{} on the TerraSentia, a small lightweight skid-steer robot from Earthsense Inc., equipped with a 6-DOF IMU, GNSS, wheel encoders, a Velodyne VLP-16 LiDAR, and a ZED 2i RGB camera. Pixel-wise semantic mask labels from the ZED stream are projected into 3D using the calibrated camera--LiDAR extrinsics and maintained in an OctoMap for traversability estimation, which is projected onto a 2D grid map for planning and risk evaluation. Onboard computation runs on a Jetson AGX Orin, with a GSM router for online GPT-4o API calls capped at 300 output tokens per query. In practice, this costs roughly \$0.28 for 200 queries (82k total input/output tokens) across multiple runs.

\subsection{Comparative Navigation in Challenging Environments}

To analyze the capability of our model, we compare \method{} with other baselines for navigation in four challenging real-world environments. The four scenarios are deliberately chosen to span distinct traversability challenges, ranging from dense geometric clutter to semantically deceptive terrain that appears flat yet cannot be traversed. We evaluate each method on: 1) \textit{success rate}, reaching the goal without collision or human intervention; 2) \textit{normalized trajectory length} relative to the straight-line distance; 3) \textit{time to goal}; and 4) \textit{minimum clearance} to obstacles, a direct measure of navigation safety. All metrics except success rate are averaged over successes.

\begin{description}[style=unboxed, leftmargin=0pt, font=\bfseries, itemsep=1pt, topsep=2pt, parsep=0pt]
\item[DWA~\cite{dwa}:] A classical dynamic window approach that uses a height map to detect obstacles and negative drops relative to the robot height for collision avoidance and goal reaching.
\item[NoMaD~\cite{sridhar2023nomad}:] A visual navigation foundation model with a transformer-diffusion policy trained on diverse robot datasets for goal-conditioned and goal-free navigation.
\item[CoNVOI~\cite{sathyamoorthy2024convoi}:] A VLM-based method that generates context-aware trajectories by identifying the robot's environment and implicit and explicit navigation behaviors.
\end{description}

\method{} succeeds in $90$--$100\%$ of runs in all four scenarios, while no baseline exceeds $60\%$ success in any of them (Table~\ref{tab:nav_runs}). The gap follows directly from how \method{} grounds traversability. Rather than reacting to instantaneous geometry or image cues, it fuses repeated VLM observations into a persistent, uncertainty-aware posterior, plans globally against worst-case cost through the CVaR objective, and slows in under-observed regions through the epistemic-uncertainty term. \method{} can therefore avoid hazards with no geometric signature while holding the highest minimum clearance in every scenario, up to $0.86$~m, trading path length for margin. Statistical analysis confirms these advantages. Pooled across all 40 trials per method, \method{}'s success rate (Wilson $95\%$ CI [87.1,99.6]) is significantly higher than every baseline under Fisher's exact test: DWA ($p=1.7 \times 10^{-10}$), CoNVOI ($p=2.4 \times 10^{-12}$) and NoMaD ($p=2.2 \times 10^{-8}$). Mann-Whitney U tests on the continuous metrics show that \method{} maintains significantly higher minimum clearance than every baseline in all pairwise scenario comparisons with $p<0.05$ after Holm-Bonferroni correction.

\begin{figure*}[!t]
    \centering
    \includegraphics[trim={0cm 1.5cm 0cm 1.5cm},clip,width=1.0\linewidth]{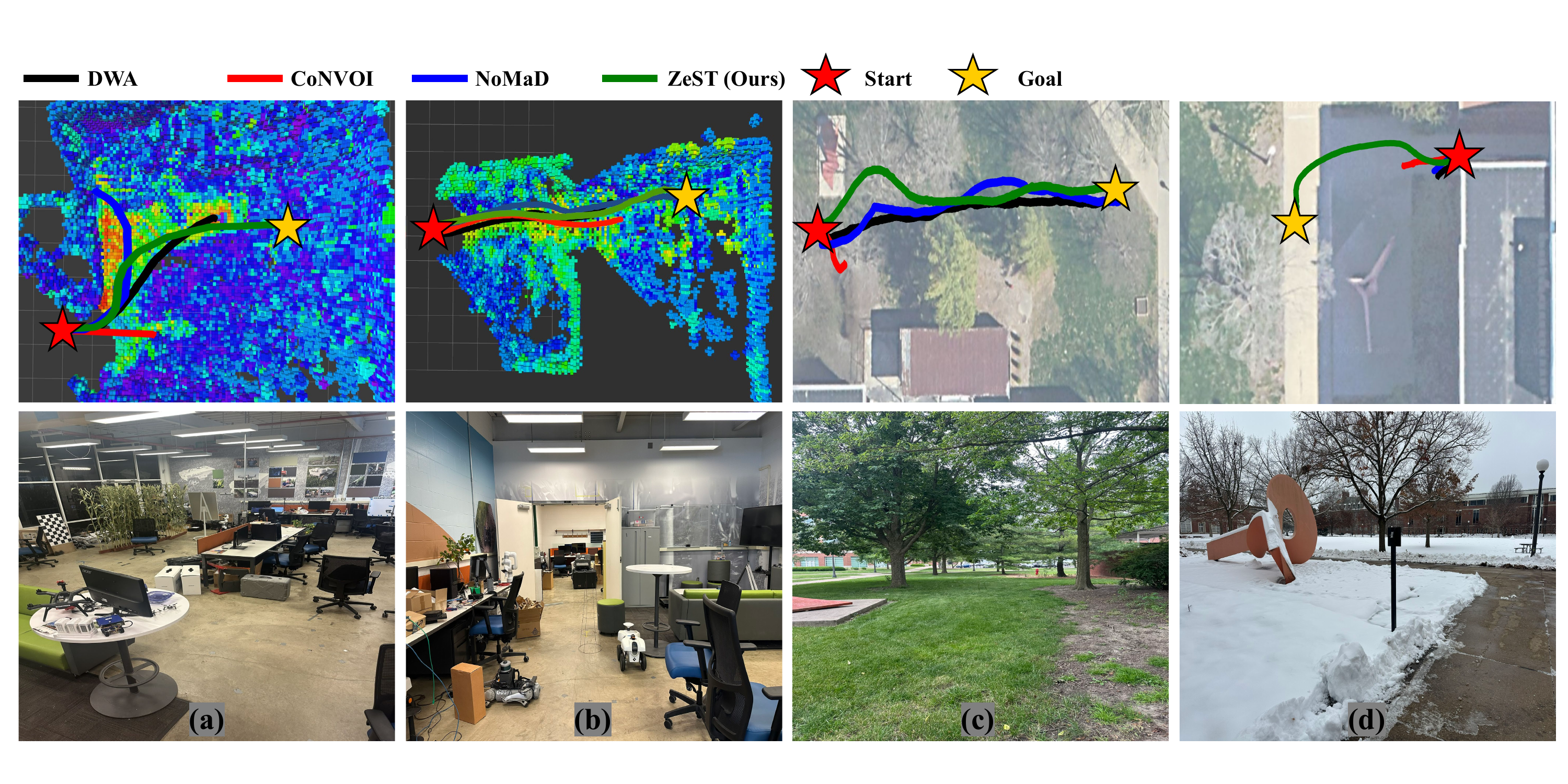}
    \caption{\textbf{Qualitative Navigation Results.} Navigation results for DWA, CoNVOI, NoMaD, and \method{} across four cluttered indoor and unstructured outdoor scenarios. (Top) \textbf{Trajectories:} Each method's executed path from one representative run is overlaid, with red and yellow stars marking the start and goal locations, respectively; indoor paths (\textbf{a},\,\textbf{b}) are shown over the traversability OctoMap generated by \method{}, where warmer regions are more traversable, and outdoor paths (\textbf{c},\,\textbf{d}) over top-down views. (Bottom) \textbf{Environments.} Scene observations: (\textbf{a}) an indoor room of desks, chairs, and fragile objects whose goal is reachable only through a single passage; (\textbf{b}) a tight corridor with furniture, wire mesh, boxes, other robots, and a thin rod on the floor; (\textbf{c}) outdoor terrain mixing grass, uneven soil, exposed roots, and bushes; and (\textbf{d}) a snow region that appears flat yet is non-traversable for the robot.}
    \label{fig:zest_exps}
    \vspace{-0.1in}
\end{figure*}

\begin{table}[t]
    \centering
    \tiny
    \setlength{\tabcolsep}{4pt}
    \caption{Comparative navigation results over $10$ runs}
    \label{tab:nav_runs}
    \begin{tabular}{clcccc}
        \toprule
        & \textbf{Method} & \makecell{\textbf{Success}\\ \textbf{Rate (\%)}~↑} & \makecell{\textbf{Norm. Traj.}\\ \textbf{Length}~(→1)} & \makecell{\textbf{Time to}\\ \textbf{Goal (s)}~↓} & \makecell{\textbf{Min. Clearance}\\ \textbf{(m)}~↑} \\
        \midrule
        \multirow{4}{*}{\textbf{1}}
        & DWA~\cite{dwa} & 40 & 1.053 $\pm$ 0.126 & 50.6 $\pm$ 1.7 & 0.41 $\pm$ 0.029 \\
        & CoNVOI~\cite{sathyamoorthy2024convoi} & 30 & 1.061 $\pm$ 0.143 & 64.9 $\pm$ 3.9 & 0.33 $\pm$ 0.056 \\
        & NoMaD~\cite{sridhar2023nomad} & 50 & 1.202 $\pm$ 0.178 & 58.5 $\pm$ 2.1 & 0.48 $\pm$ 0.031 \\
        & \method{}~(Ours) & 100 & 1.136 $\pm$ 0.029 & 61.6 $\pm$ 1.8 & 0.53 $\pm$ 0.059 \\
        \cmidrule(lr){2-6}
        \multirow{4}{*}{\textbf{2}}
        & DWA~\cite{dwa} & 30 & 1.012 $\pm$ 0.041 & 58.7 $\pm$ 2.1 & 0.23 $\pm$ 0.051 \\
        & CoNVOI~\cite{sathyamoorthy2024convoi} & 20 & 1.081 $\pm$ 0.072 & 60.9 $\pm$ 5.4 & 0.30 $\pm$ 0.031 \\
        & NoMaD~\cite{sridhar2023nomad} & 60 & 1.029 $\pm$ 0.065 & 54.9 $\pm$ 2.8 & 0.25 $\pm$ 0.022 \\
        & \method{}~(Ours) & 100 & 1.101 $\pm$ 0.056 & 58.4 $\pm$ 3.5 & 0.44 $\pm$ 0.060 \\
        \cmidrule(lr){2-6}
        \multirow{4}{*}{\textbf{3}}
        & DWA~\cite{dwa} & 60 & 1.192 $\pm$ 0.169 & 201.2 $\pm$ 12.8 & 0.50 $\pm$ 0.031 \\
        & CoNVOI~\cite{sathyamoorthy2024convoi} & 40 & 1.127 $\pm$ 0.125 & 236.5 $\pm$ 11.2 & 0.49 $\pm$ 0.029 \\
        & NoMaD~\cite{sridhar2023nomad} & 60 & 1.158 $\pm$ 0.267 & 208.7 $\pm$ 10.7 & 0.51 $\pm$ 0.049 \\
        & \method{}~(Ours) & 100 & 1.339 $\pm$ 0.168 & 226.6 $\pm$ 9.8 & 0.80 $\pm$ 0.085 \\
        \cmidrule(lr){2-6}
        \multirow{4}{*}{\textbf{4}}
        & DWA~\cite{dwa} & 0 & -- & -- & -- \\
        & CoNVOI~\cite{sathyamoorthy2024convoi} & 10 & 1.246 & 85.1 & 0.59 \\
        & NoMaD~\cite{sridhar2023nomad} & 0 & -- & -- & -- \\
        & \method{}~(Ours) & 90 & 1.468 $\pm$ 0.212 & 80.3 $\pm$ 5.3 & 0.86 $\pm$ 0.076 \\
        \bottomrule
    \end{tabular}
    \vspace{-0.2in}
\end{table}

\textbf{Scenario 1} is a cluttered indoor space whose goal is occluded and reachable only through a single passage. \method{} routes through the most traversable regions of its map (Fig.~\ref{fig:zest_exps}), accepting a longer path for the largest clearance. DWA takes a shorter geometric path but repeatedly collides with clutter and needs intervention. NoMaD, run zero-shot, wanders and generalizes poorly to the scene. CoNVOI re-plans locally at each step, commits to momentarily clear directions, and cannot recover, also posting the longest time to goal as its viewpoint shifts while the robot turns. \textbf{Scenario 2} is a tight corridor whose main hazards are a thin rod and a wire mesh. \method{} detects both and maneuvers around them at the highest clearance and slows to preserve recovery time. DWA and NoMaD miss the rod and drive over it, and neither holds clearance from the wire mesh or box. CoNVOI avoids the rod at first, but its local-only planning leaves no room to recover.

\textbf{Scenario 3} is an outdoor scene with a straight-ahead goal across grass, uneven soil, exposed roots, and bushes. This terrain is the most forgiving of the four, so the baselines reach the goal more often, with DWA and NoMaD at $60\%$ and CoNVOI at $40\%$. The main separation is safety: \method{} marks the uneven soil as low-traversability and takes a longer grass route, maintaining $0.80$~m clearance, while the baselines cut across rough ground and keep only about $0.50$~m near roots and bushes. \textbf{Scenario 4} places the goal beyond snow that appears flat but can immobilize a small wheeled robot, creating a semantic hazard with no geometric signature. \method{} identifies the snow as non-traversable and routes along exposed concrete, even though snow is never named in the prompt. DWA and NoMaD treat the snow as free space and drive into it; CoNVOI finds the concrete but stays too close to the snow edge.

\begin{table}[!tb]
    \centering
    \tiny
    \setlength{\tabcolsep}{4pt}
    \caption{Component ablation results}
    \label{tab:ablation}
    \begin{tabular}{clcccc}
        \toprule
        & \textbf{Variant} & \makecell{\textbf{Success}\\ \textbf{Rate (\%)}~↑}
        & \makecell{\textbf{Norm. Traj.}\\ \textbf{Length}~(→1)}
        & \makecell{\textbf{Time to}\\ \textbf{Goal (s)}~↓}
        & \makecell{\textbf{Min. Clearance}\\ \textbf{(m)}~↑} \\
        \midrule
        \multirow{4}{*}{\rotatebox[origin=c]{90}{\textbf{Indoor}}}
        & \method{}~(Full) & 80        & 1.546 $\pm$ 0.254 & 56.1 $\pm$ 2.7 & 0.32 $\pm$ 0.154 \\
        & w/o Fusion       & 30        & 1.665 $\pm$ 0.301 & 54.9 $\pm$ 3.3 & 0.22 $\pm$ 0.129 \\
        & w/o Risk         & 10        & 1.271             & 51.1           & 0.23 \\
        & Raw Oracle       & 10        & 1.031             & 46.6           & 1.42 \\
        \cmidrule(lr){2-6}
        \multirow{4}{*}{\rotatebox[origin=c]{90}{\textbf{Outdoor}}}
        & \method{}~(Full) & 90             & 1.509 $\pm$ 0.159 & 62.3 $\pm$ 2.7 & 0.90 $\pm$ 0.074 \\
        & w/o Fusion       & 30             & 1.438 $\pm$ 0.027 & 64.6 $\pm$ 1.4 & 0.76 $\pm$ 0.321 \\
        & w/o Risk         & \phantom{0}0   & --                & --             & -- \\
        & Raw Oracle       & 30             & 1.204 $\pm$ 0.462 & 54.9 $\pm$ 9.3 & 0.67 $\pm$ 0.250 \\
        \bottomrule
    \end{tabular}
    \vspace{-0.2in}
\end{table}

\subsection{Ablation Study}

To measure the contribution of each component, we ablate fusion and the risk-aware cost, the two steps that turn the zero-shot VLM signal into a navigation decision, on one indoor and one outdoor scenario over $10$ runs each under the same protocol and metrics as Table~\ref{tab:nav_runs}. The ablation uses paired indoor and outdoor courses: the indoor course contains narrow and semantically ambiguous hazards, including a thin rod, chairs, fire extinguishers, glass, a trash bin, and a door laid horizontally on the floor, while the outdoor course adds trees, vine trees, benches, and scattered clutter. Without fusion, the per-cell Bayesian posterior is replaced by a running mean of the VLM scores. When the risk layer is removed, fusion is retained, but planning relies on posterior-mean traversability rather than the risk-aware CVaR cost. In the Raw Oracle setting, both components are disabled, and planning proceeds directly from the raw per-query measurements.

Without fusion, success falls to $30\%$ and successful indoor runs are longer with lower clearance than the full system. Without the risk layer, success drops to $10\%$ indoors, and $0\%$ outdoors, so trajectory metrics are based on too few runs to indicate stable behavior. Raw Oracle can produce short or high-clearance successful runs, but low success rates show that raw per-query measurements are not enough for reliable navigation. The full system is the only variant to achieve high success in both environments.

\begin{figure*}[!t]
    \centering
    \includegraphics[width=0.78\textwidth]{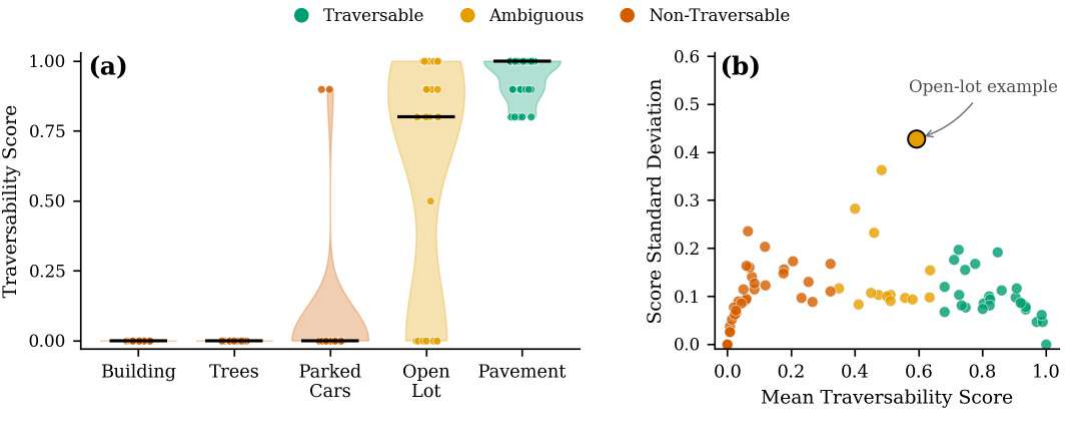}
    \caption{\textbf{VLM Query Variability.} Traversability scores from repeated VLM queries on fixed regions and prompts. \textbf{(a)} Per-region scores from $30$ queries on one held-out frame, grouped by terrain type; black bars mark medians. Scores stay tight and repeatable on confident terrain like pavement, trees, and buildings, and spread widely on ambiguous terrain like the open lot. \textbf{(b)} Per-region score standard deviation versus mean score across $83$ regions from $17$ frames; the outlined marker is the open-lot region from (a). The spread is lowest at both ends of the score range and peaks at intermediate scores, so the per-query variability concentrates on genuinely ambiguous terrain rather than scattering at random.}
    \label{fig:query_variance}
\end{figure*}

\subsection{Latency Robustness}
\label{sec:latency}

In our system, GPT-4o queries average $1.82$~s, with most responses in the range $1$--$2.5$~s and occasional outliers near $5$~s. Such second-scale variable delays are consistent with recent robotics systems that place VLM reasoning outside high-rate control loops or query foundation models intermittently~\cite{hirose2026asyncvla, chen2025commonsense}. To absorb them, \method{} plans on an OctoMap with a $10$-meter horizon that retains the accumulated traversability posterior between queries. At the same time, an epistemic-uncertainty cost in the MPPI controller slows the robot in under-observed regions, so it keeps navigating on past observations while a new VLM response is pending.

We test latency tolerance directly on a run set independent of Table~\ref{tab:nav_runs}, adding fixed delays of $1$--$4$~s into the querying pipeline, raising the effective query latency to as much as $6.1$~s. As shown in Table~\ref{tab:latency_exp}, success holds at $100\%$ ($95\%$ CI $[72,100]$) up to an average latency of $3.6$~s ($+2$~s), and degrades only past $5$~s, where success falls to $80\%~[49,94]$ and $70\%~[40,89]$, outside measured GPT-4o range. Average speed declines from $61\%$ to $38\%$ of nominal as latency grows, consistent with the uncertainty-driven slowdown. Success rates are over $10$ runs; time to goal and average speed are over successful runs.

\subsection{Zero-Shot vs. In-Domain Learned Traversability Analysis}

The probabilistic fusion and risk-aware layers in \method{} only help if each zero-shot VLM measurement carries a usable traversability signal. We test this directly by evaluating, before fusion, how discriminative one zero-shot VLM measurement is relative to a learned traversability sensor operating under familiar sensing and terrain conditions. We compare zero-shot VLM-derived scores against WayFAST~\cite{gasparino2022wayfast}, trained on TerraSentia data, on a held-out outdoor traverse collected with the same robot platform. WayFAST runs on native RGB-D input, and both sensors are scored on identical image regions; \method{} receives no in-domain training or adaptation.

\begin{table}[!tb]
    \centering
    \caption{Injected VLM query delays in Indoor Scenario 2}
    \label{tab:latency_exp}
    \setlength{\tabcolsep}{4pt}
    \begin{tabular}{ccccc}
    \toprule
        \makecell{\textbf{Added}\\ \textbf{Delay (s)}} & \makecell{\textbf{Avg.}\\ \textbf{Latency (s)}} & \makecell{\textbf{Success}\\ \textbf{Rate (\%)}~↑} & \makecell{\textbf{Time to}\\ \textbf{Goal (s)}~↓} & \makecell{\textbf{Avg. Speed}\\ \textbf{(\% nominal)}} \\
    \midrule
         None  & 1.82 & 100 & 55.3\,$\pm$\,2.3 & 61 \\
         $+1.0$ & 2.81 & 100 & 57.8\,$\pm$\,1.8 & 57 \\
         $+2.0$ & 3.63 & 100 & 61.8\,$\pm$\,2.4 & 49 \\
         $+3.0$ & 5.15 & \phantom{0}80 & 64.7\,$\pm$\,2.3 & 45 \\
         $+4.0$ & 6.12 & \phantom{0}70 & 72.4\,$\pm$\,2.7 & 38 \\
    \bottomrule
    \end{tabular}
    \vspace{-0.1in}
\end{table}

We ground the comparison in human judgment using the human disagreement rate (HDR), introduced by W-RIZZ~\cite{schreiber2024wrizz} for relative traversability evaluation from human-annotated pairs. We adapt HDR from pixels to SLIC regions, so each annotation compares two terrain segments. Across the held-out traverse, two annotators label $400$ sampled region pairs by selecting the more traversable segment or marking a tie, with strong agreement (Krippendorff $\alpha_{\mathrm{ord}}=0.80$). HDR is computed on the $318$ clear-preference pairs and measures the fraction of model preferences disagreeing with the annotators at the score-gap threshold $\tau$. Lower HDR indicates stronger agreement with human labels, as reported in Table~\ref{tab:hdr_main}. A model registers a preference only when its score gap exceeds $\tau$; smaller gaps are recorded as ties, which count as disagreements on clear-preference pairs. Increasing $\tau$ therefore probes the usable dynamic range of a sensor's scores.

At the tightest threshold, \method{} and WayFAST are comparable because small score gaps can express preferences. As $\tau$ increases, the evaluation emphasizes confident distinctions between safer and hazardous terrain, and \method{} maintains lower HDR than WayFAST. Spearman correlation gives a slightly different view: WayFAST has a marginally higher rank correlation ($0.62$ versus $0.58$ for the VLM), but Spearman rewards monotone ordering regardless of score magnitude, whereas HDR requires a gap above $\tau$. WayFAST can therefore rank regions acceptably while still collapsing to ties: its scores concentrate near a single value and produce over $99\%$ ties at $\tau=0.5$. Before fusion, the threshold trend shows that a single VLM measurement carries a usable traversability signal.

To measure single-query stability, we re-query the deployment VLM on frozen inputs. For a $17$-frame subset of the held-out traverse, we reuse the same numbered SLIC image and \method{} prompt, issuing $30$ repeated queries with the same deployment sampling settings. The $83$ evaluated regions carry the same region indices as the HDR analysis. Malformed or refused replies are discarded before scoring ($19.6\%$ of queries), without realigning the region indices.

Table~\ref{tab:hdr_main} shows useful traversability signal in one VLM measurement, while Fig.~\ref{fig:query_variance} explains why one query should not be trusted alone. Across the $83$ regions, the per-query standard deviation has a median of $0.094$ (IQR $[0.062,0.125]$) and reaches $0.428$. By turning repeated-query variance into uncertainty, the NIG posterior shows that the zero-shot signal is discriminative but unreliable without fusion.

\begin{table}[!tb]
    \centering
    \caption{Human disagreement rate for zero-shot\\ and in-domain traversability scores}
    \label{tab:hdr_main}
    \setlength{\tabcolsep}{4pt}
    \begin{tabular}{lccc}
    \toprule
        \textbf{Method} & \textbf{HDR@0.1}~↓ & \textbf{HDR@0.25}~↓ & \textbf{HDR@0.5}~↓ \\
    \midrule
        \method{} & $0.32\,[0.22,0.41]$ & $0.31\,[0.21,0.40]$ & $0.37\,[0.27,0.47]$ \\
        WayFAST & $0.29\,[0.23,0.35]$ & $0.48\,[0.42,0.55]$ & $0.71\,[0.66,0.78]$ \\
    \bottomrule
    \end{tabular}
    \vspace{-0.1in}
\end{table}

\section{Conclusion}

In this work, we presented \method{}, a navigation method which treats a zero-shot VLM as a stochastic traversability estimator and fuses repeated measurements into an NIG posterior that separates aleatoric and epistemic uncertainty. A CVaR-based RRT* planner enables global path planning for long-term autonomous navigation, while a short-term epistemic-uncertainty-aware sampling-based model predictive controller (MPC) is used for fast and reactive local planning. As demonstrated, \method{} improves navigation safety by fusing VLM predictions. The proposed pipeline achieves better navigational performance compared to learned and classical baselines while requiring no in-domain training or adaptation. However, repeated NIG updates cannot correct systematic VLM errors and may instead reinforce incorrect estimates. Incorporating independent feedback, such as proprioception, could help detect systematic errors.

\addtolength{\textheight}{-1cm}   

\bibliographystyle{IEEEtran}
\bibliography{references}

\end{document}